\documentclass[letterpaper, 10 pt, conference]{ieeeconf}

\IEEEoverridecommandlockouts                              
\usepackage[pdftex]{graphicx}
\usepackage{multirow}
\usepackage{url}
\usepackage{hyperref}
\usepackage{cite}

\usepackage{amsthm}
\usepackage{amsmath}
\usepackage{amsfonts}
\usepackage[OT1]{fontenc} 
\usepackage{bm}
\usepackage{centernot}
\usepackage{algorithm}
\usepackage{algpseudocode}

\usepackage[nonumberlist, nogroupskip, section=section, numberedsection=autolabel]{glossaries}
\newacronym{nlp}{NLP}{Nonlinear Program}
\newacronym{to}{TO}{Trajectory Optimization}
\newacronym{ocp}{OCP}{Optimal Control Problem}
\newacronym{pho}{P-HO}{Probabilistic Homotopy Optimization}
\newacronym{rho}{R-HO}{RRT Homotopy Optimization}
\newacronym{lho}{LI-HO}{Linear Interpolation Homotopy Optimization}
\newacronym{prm}{PRM}{Probabilistic Roadmap}
\newacronym{rrt}{RRT}{Rapidly Exploring Random Tree}

\newtheorem{definition}{Definition}

\algnewcommand\algorithmicinput{\textbf{Hyperparams:}}
\algnewcommand\Hyperparameters{\item[\algorithmicinput]}

\usepackage{color}

\newcommand{\R}{\mathbb{R}}

\newcommand{\variable}{\mathbf{x}}
\newcommand{\localMin}{\variable^{*}}
\newcommand{\variableSpace}{\mathcal{X}}
\newcommand{\feasibleSet}{\mathcal{C}}
\newcommand{\NLP}[1]{\mathcal{P}_{#1}}
\newcommand{\param}{\bm{\theta}}

\newcommand{\paramSpace}{\mathcal{Q}} 
\newcommand{\easyParam}{\param_0}
\newcommand{\goalParam}{\param^*}

\newcommand{\solve}[2]{\mathrm{solve}\left({\NLP{#1}},#2\right)}
\newcommand{\Basin}[2]{\mathcal{B}_{#1}\left(#2\right)}

\newcommand{\hparam}{{\bm{\lambda}}}
\newcommand{\hparamSize}{d}
\newcommand{\hparamSpace}[1]{\mathcal{H}_{#1}}
\newcommand{\hparamSeq}{\bm{\Lambda}}
\newcommand{\SolutionCurve}{{\mathcal{M}_S}}
\newcommand{\paramMap}[1]{\Gamma\left(#1\right)}
\newcommand{\Feasy}{F_{easy}}

\newcommand{\tree}{\tau}
\newcommand{\Node}[2]{\mathbf{N}_{#1,#2}}

\newcommand{\closeNode}{\mathbf{N}_{close}}
\newcommand{\NodeExpand}[2]{\Node{#1}{#2}=\left(\localMin_{#1},\hparam_{#2}\right)}

\newcommand{\trials}{\mathcal{S}_\mathcal{A}}

\newcommand{\hparamset}{\mathcal{S}_{\hparam}}
\newcommand{\goalProb}{P_g}
\newcommand{\trialDensity}{\rho_\mathcal{A}}

\newcommand{\timeLimit}{T_{max}}
\newcommand{\numExperiments}{30}

\title{\LARGE \bf Probabilistic Homotopy Optimization for Dynamic Motion Planning}

\author{Shayan Pardis\footnotemark$^{*1}$ Matthew Chignoli\footnotemark$^{*2}$ and Sangbae Kim\footnotemark$^{2}$
\thanks{$^{*}$Equal contribution.}
\thanks{$^{1}$Department of Electrical Engineering and Computer Science and $^{2}$Department of Mechanical Engineering, Massachusetts Institute of Technology, Cambridge, MA 02139, USA: {\tt\small chignoli@mit.edu}}}%

\begin{document}

\maketitle
\thispagestyle{empty} 
\pagestyle{empty}

\begin{abstract}
We present a homotopic approach to solving challenging, optimization-based motion planning problems.
The approach uses Homotopy Optimization, which, unlike standard continuation methods for solving homotopy problems, solves a sequence of constrained optimization problems rather than a sequence of nonlinear systems of equations.
The insight behind our proposed algorithm is formulating the discovery of this sequence of optimization problems as a search problem in a multidimensional homotopy parameter space.
Our proposed algorithm, the Probabilistic Homotopy Optimization algorithm, switches between solve and sample phases, using solutions to easy problems as initial guesses to more challenging problems.
We analyze how our algorithm performs in the presence of common challenges to homotopy methods, such as bifurcation, folding, and disconnectedness of the homotopy solution manifold.
Finally, we demonstrate its utility via a case study on two dynamic motion planning problems: the cart-pole and the MIT Humanoid.
\end{abstract}


\section{Introduction}
\begin{figure}[thb]
    \centering
    \includegraphics[width=2.75in]{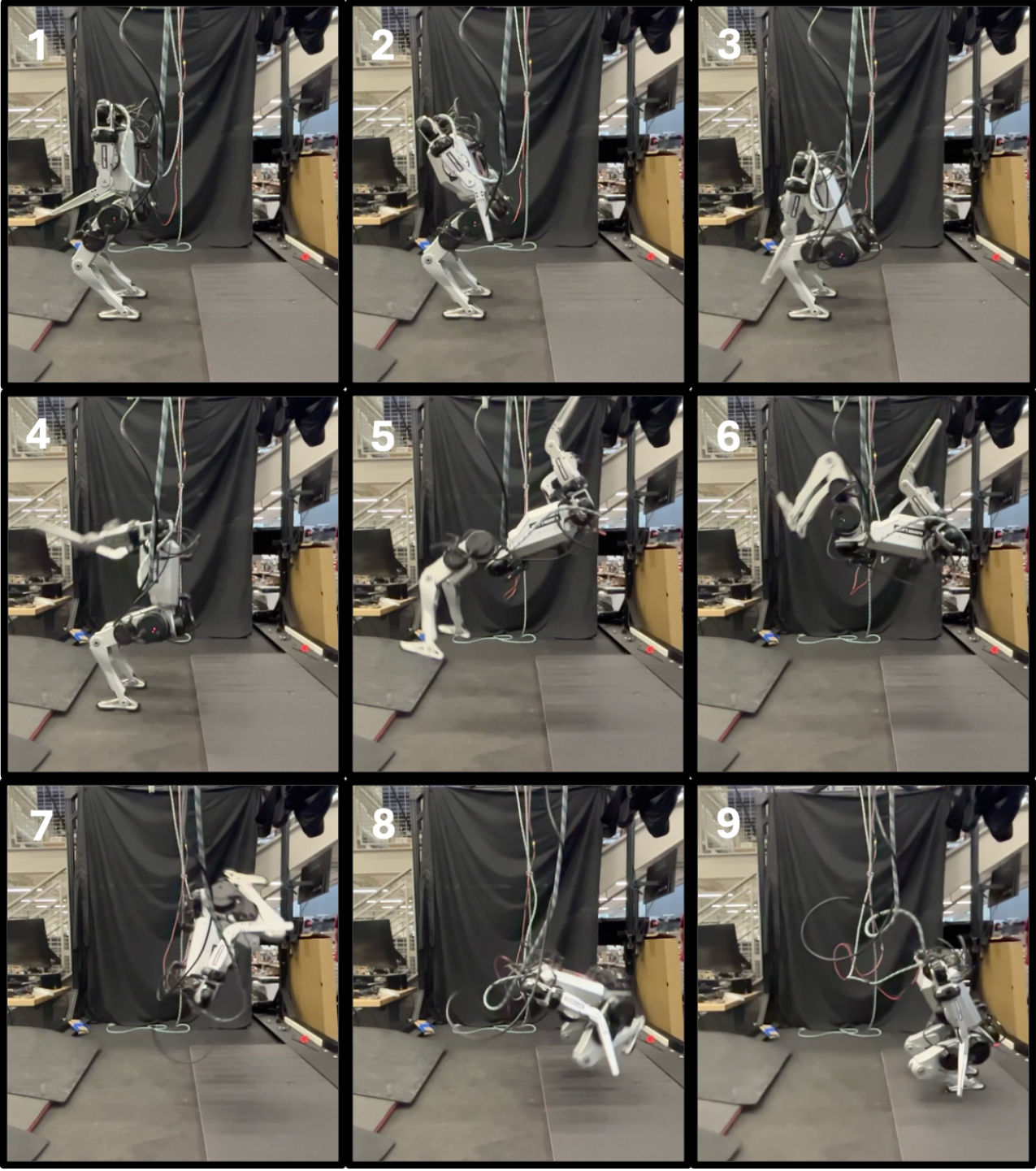}
    \caption{Framewise animation of the back flip motion planned using Probabilstic Homotopy Optimization.}
    \label{fig:front_flip}
\end{figure}

The ability of optimization-based motion planners to generalize to unseen situations and reason about challenging constraints makes them powerful tools.
Highly efficient motion planners can run in real-time for MPC~\cite{grandia2023perceptive,kim2023contact}, while less efficient motion planners are still valuable as sources of high-quality reference data for training reinforcement learning policies~\cite{jenelten2024dtc, brakel2022learning, marcucci2023motion}.
Highly dynamic motions such as the back flip shown in Fig.~\ref{fig:front_flip} often have a highly non-convex cost landscape~\cite{bledt2020extracting}, which makes it difficult to converge to a solution reliably.
This work introduces a systematic approach to solving such challenging nonlinear trajectory optimization problems. 
Our approach involves defining a homotopy between simple and challenging motion planning problems and then
probabilistically searching the potentially multidimensional homotopy parameter space in a manner robust to common pitfalls such as bifurcation and disjoint solution manifolds.
We demonstrate our method's ability to solve challenging nonlinear optimization problems that are not solvable with off-the-shelf globalization strategies.

Homotopy methods~\cite{nocedal1999numerical} are a common approach to solving challenging Nonlinear Programs such as those arising from robotic motion planning.
The Interior Point Method upon which many modern \gls{nlp} solvers are built~\cite{wachter2006implementation,byrd2006k} involves a homotopy over the nonlinear system of equations corresponding to the optimality conditions, where the homotopy parameter encodes a relaxation of the problem constraints.
While this method has proven to be very effective for solving a wide variety of challenging Nonlinear Programs~\cite{wachter2006implementation,byrd2006k}, restricting the homotopy to the optimality relaxation parameter does not allow domain knowledge about the problem or its structure to be leveraged.
The Homotopy Optimization Method~\cite{Dunlavy2005HomotopyOM} can exploit such domain knowledge because it involves solving a sequence of constrained minimization problems rather than systems of nonlinear equations.
Versions of the Homotopy Optimization Method have been employed for robotic motion planning~\cite{rosa2014extending,raff2022generating,jenelten2022tamols,turski2023staged}, but have always pre-specified the full sequence of optimizations to solve.
In contrast, our proposed algorithm eliminates the need for pre-specification of the sequence.

While the Homotopy Optimization Method is more general than homotopy over optimality relaxation, it can lead to a more challenging homotopic landscape due to bifurcations, folds, abbreviated paths, and disconnected branches~\cite{watson1989globally}.
Adding homotopy parameters can alleviate the folding problem~\cite{wolf1996multiparameter}.
However, adding homotopy parameters necessitates advanced methods for tracing a solution in the now multidimensional homotopy parameter space~\cite{vazquez2011powering, grenat2019multi}.
These multidimensional tracing methods require gradients of the homotopy map with respect to the problem variables and homotopy parameters, which are not readily available in Homotopy Optimization.

In this work, we propose \gls{pho} algorithm for solving challenging \gls{nlp}s corresponding to dynamic motion planning for robotic systems.
Like the stochastic extension of the Homotopy Optimization Method~\cite{Dunlavy2005HomotopyOM}, our approach involves stochastically selecting and then solving a sequence of minimization problems ranging from easy to difficult.
However, we offer a generalized formulation that (i) enables more flexible homotopy definitions, (ii) requires less hyperparameter tuning, and (iii) improves explorations to find better local minima.
Additionally, our approach does not require gradients of the homotopy map to trace a solution.
The crucial features of our homotopy method, and thus the main contributions of the paper, are
\begin{itemize}
    \item Formulating the traversal of the homotopy's Solution Manifold as a search problem
    \item A novel sampling-based algorithm to efficiently traverse the Solution Manifold
    \item Support for multidimensional homotopy parameterizations
\end{itemize}

\gls{pho}'s method of growing the tree eliminates the need to tune hyparameters related to the pre-specification the homotopy path.
Furthermore, for each parameterization in homotopy space, \gls{pho} solves the problem multiple times with different initial guesses, so it can explore multiple local minimas and choose the best among them.

Lastly, the ability to define multidimensional homotopy parameters allows the algorithm, rather than the user, to find the rate at which the difficulty of different aspects of the problem is increased.
We demonstrate the capabilities of our proposed homotopy method by implementing it to solve a cart pole swing-up problem and multiple highly dynamic motion planning problems involving the MIT Humanoid robot.

The rest of the paper is organized as follows.
Background information on relevant mathematical concepts and notation is provided in Sec.~\ref{sec:background}.
Sec.~\ref{sec:pho} formalizes our problem statement, describes the \gls{pho} algorithm, and analyzes its properties.
For comparison, two alternatives to the \gls{pho} algorithm are presented in Sec.~\ref{sec:liho_and_rho}.
Sec.~\ref{sec:results} presents the results from case studies using the proposed algorithms to solve dynamic motion planning problems, and Sec.~\ref{sec:conclusion} summarizes and concludes the work.
\section{Background} \label{sec:background}

\subsection{Trajectory Optimization}
Nonlinear programming involves solving a problem of the form:
\begin{equation}
\begin{split}
    \NLP{\param} = \min_{\variable\in\variableSpace} \quad &f(\variable, \param) \\
    \text{s.t.} \quad& \mathbf{c}(\variable, \param) = \mathbf{0} \\
    &\variable \ge \mathbf{0}
\end{split} \label{eqn:parameterized_nlp}
\end{equation}
where $\variable \in \variableSpace$ are the decision variables, $f:\variableSpace\rightarrow\R$ is an objective function, $\mathbf{c}:\variableSpace\rightarrow\R^m$ is a set of equality constraints, and $\param\in\paramSpace$ is a set of parameters whose value is fixed prior to solving.
The feasible set of decision variables $\feasibleSet\subset\variableSpace$ contains all decision variables satisfying $\mathbf{c}(\variable, \param) = \mathbf{0}$ and $\variable \ge \mathbf{0}$.
Any nonlinear program can be converted to the form of~\eqref{eqn:parameterized_nlp} through slack variables~\cite{fletcher2000practical}.

In trajectory optimization, the goal is to find the state and control trajectories that optimally accomplish a desired task.
Thus, the decision variables encode these state and control trajectories, the cost encodes what it means to ``optimally" accomplish the task, and the constraints encode the dynamics of the system as well as any other task-specific path or boundary constraints.

Analytical solutions for such trajectory optimization problems are typically intractable.
Thus, due to the ready availability of local gradient information, gradient-based numerical methods such as Sequential Quadratic Programming~\cite{nocedal1999numerical}, Interior Point Methods~\cite{nocedal1999numerical}, and Differential Dynamic Programming~\cite{mayne1966second} are used to solve these problems.
We define the following operator to represent solving~\eqref{eqn:parameterized_nlp} with a numerical solver:
\begin{equation}
    \solve{\param}{\variable} : \variableSpace\rightarrow\variableSpace,
\end{equation}
where $\NLP{\param}$ is the NLP being solved and $\variable$ is the initial point used by the solver.
The solver will return a locally optimal solution $\localMin\in\feasibleSet$ if and only if the initial point is in the basin of attraction of $\localMin$.
Otherwise, it will return an unusable, infeasible solution not in $\feasibleSet$.
\begin{definition}[Basin of Attraction] The set of all initial guesses $\variable$ for which the numerical solver finds a locally optimal solution $\localMin$ to the problem $\NLP{\param}$.
    \[
    \Basin{\param}{\localMin} = \{ \variable : \solve{\param}{\variable} = \localMin \}.
    \]
\end{definition}
Note that the behavior of $\solve{\param}{\cdot}$ and the Basin of Attraction will depend on the solver used. 
Our proposed approach will work for any solver, but results will vary.

\subsection{Homotopy Methods}

Homotopy methods are numerical techniques that aim to solve a challenging nonlinear problem $F(\variable)$ by gradually transforming an easy problem $\Feasy(\variable)$ into $F(\variable)$ via a homotopy~\cite{nocedal1999numerical}.
The homotopy defines a continuous relationship between $\Feasy(\variable)$ and $F(\variable)$ as a function of the homotopy parameter $\hparam\in\hparamSpace{\hparamSize}$, where
\[
\hparamSpace{\hparamSize} = \left\{\hparam=\begin{bmatrix} \lambda_1 \\ \vdots \\ \lambda_d \end{bmatrix} :  0 \le \lambda_i \le 1 \, \forall i=1,\hdots,\hparamSize \right\}. 
\]
Typically, $\hparamSize=1$, but the homotopy parameter can be multidimensional, as will be the case in this work.
A homotopy map $H(\variable,\hparam)$ is defined such that when $\hparam=\mathbf{0}_\hparamSize$, the solution to $H(\variable,\mathbf{0}_\hparamSize)$ is easily obtained solution to $\Feasy(\variable)$, and the solution to $H(\variable,\mathbf{1}_\hparamSize)$ is the solution to $F(\variable)$.
Here, $\mathbf{0}_\hparamSize$ and $\mathbf{1}_\hparamSize$ are the $\hparamSize$-dimensional vectors of zeroes and ones, respectively.

Homotopy methods are most often used to solve systems of nonlinear equations.
However, they can also be used to solve optimization problems~\cite{Dunlavy2005HomotopyOM}.
This technique is referred to as ``Homoptopy Optimization."
In Homotopy Optimization, $\Feasy(\variable)$ and $F(\variable)$ encode \gls{nlp}s rather than systems of equations.
How difficult an NLP is to solve depends on the problem parameters $\param$.
For example, $\param=\easyParam$ might lead to simple constraints that permit a large basin of attraction, while $\param=\goalParam$
might encode challenging constraints with a small basin of attraction.
To relate problem difficulty to homotopy, we define a mapping between homotopy parameters and optimization problem parameters, 
\[
\paramMap{\hparam}:\hparamSpace{\hparamSize}\rightarrow\paramSpace,
\]
such that $\hparam=\mathbf{0}_\hparamSize$ maps to the ``easy" problem $\easyParam$ and $\hparam=\mathbf{1}_\hparamSize$ maps to the ``difficult" goal problem $\goalParam$.

Homotopy optimization methods aim to find a solution to $\hparam=\mathbf{1}_\hparamSize$ using intermediate points on the Solution Manifold.

\begin{definition}[Solution Manifold]
    The Solution Manifold is the manifold formed in $\variableSpace\times\hparamSpace{d}$ by the union of all solvable local minima $\localMin$ of $\NLP{\paramMap{\hparam}}$.
    \[
    \SolutionCurve = \{(\localMin, \hparam) : \exists \variable\in \Basin{\paramMap{\hparam}}{\localMin} \}.
    \]
\end{definition}

\section{Probabilistic Homotopy Optimization} \label{sec:pho}

In this work, we propose \gls{pho} to solve challenging trajectory optimization problems.
The approach involves searching in the space $\variableSpace\times\hparamSpace{d}$.
We sample homotopy parameters $\hparam\in\hparamSpace{d}$ and then use previous solutions to find as many points on the Solution Manifold (i.e., local minima) as possible for each sampled $\hparam$.
This sample and solve process is repeated until a solution to the goal \gls{nlp} is found.
This section will formally define our problem, detail our proposed search algorithm, and briefly analyze its properties.

\subsection{Problem Statement}
Given a problem $\NLP{\goalParam}$ and a parameter map $\paramMap{\cdot}$, the goal of a homotopy optimization method is to find a homotopy path, represented with the sequence of parameters $\hparamSeq=\{\hparam_0,\hdots,\hparam_K\}$, such that for a trivial initial guess $\variable_0$
\begin{equation}
    \begin{cases}
        \variable^*_0 = \solve{\paramMap{\hparam_{0}}}{\variable_{0}} \\
        \variable^*_k = \solve{\paramMap{\hparam_{k}}}{\variable^*_{k-1}} & \forall k = 1,\hdots,K \\
        \variable^*_{k-1}\in\Basin{\paramMap{\hparam_{k}}}{\variable^*_{k}} & \forall k = 1,\hdots,K \\
        \hparam_0 = \mathbf{0}_\hparamSize, \quad \hparam_K = \mathbf{1}_\hparamSize.
    \end{cases} \label{eqn:param_seq_requirements}
\end{equation}
The behavior of a numerical solver is, in general, intractable to characterize.
We, therefore, treat the numerical solver as a black box, making no assumptions about the structure of the Solution Manifold or the Basins of Attraction for any of the problems $\NLP{\param}$. 
Conventional path-tracing homotopy methods~\cite{nocedal1999numerical} require access to gradient information and are limited to following a single connected component of the Solution Manifold.
Furthermore, the homotopy parameter's multidimensionality makes it difficult to prescribe the relative rates of change of its components.
As such, the problem of finding the optimal sequence $\hparamSeq$ is well-suited to be solved with a sampling-based approach.
Our sampled-based approach involves growing an ``Optimization Tree" that captures information about which \gls{nlp}s have been solved.

\begin{definition}[Optimization Tree]
    An Optimization Tree $\tree$ is a \textbf{rooted} tree data structure that keeps track of the optimization problems that have been solved.
    Nodes $\Node{i}{j}$ of the tree correspond to pairs of local minima and the problem parameters that lead to them: $\NodeExpand{i}{j}$.
    An edge from $\Node{i}{j}$ to $\Node{k}{\ell}$ denotes that $\variable^*_i\in\Basin{\paramMap{\hparam_\ell}}{\variable_k^*}$.
\end{definition}

The root node of the tree is $\Node{0}{0} = (\localMin_0,\mathbf{0}_\hparamSize)$ and is assumed to have a Basin of attraction such that it is easy to initialize the tree with a trivial initial guess $\variable_{0}$.
Thus, finding a sequence of parameters $\hparamSeq$ satisfying~\eqref{eqn:param_seq_requirements} is equivalent to
finding a path from the root $\Node{0}{0}$ to a node $\Node{i}{K}$ where $\hparam_K=\mathbf{1}_\hparamSize$. 
Different branches of the tree allow for searching the parameter space in different homotopic directions.

\subsection{Probabilistic Homotopy Optimization}
\begin{figure}
      \centering
      \includegraphics[width=1.0\linewidth]{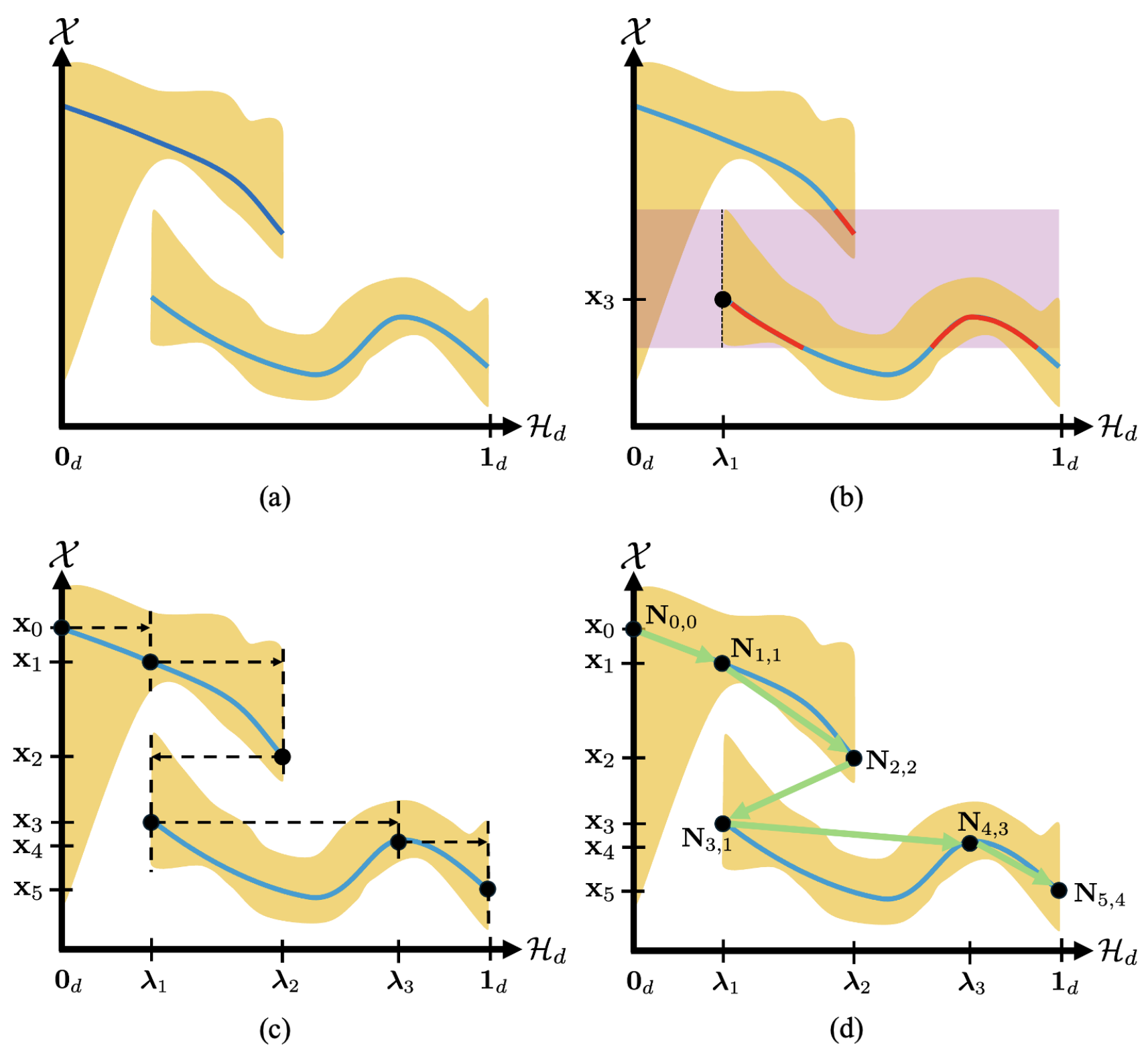}
      \caption{Illustration of the \gls{pho} algorithm: 
      (a) Solution Manifold $\SolutionCurve$ (blue) and the associated Basin of Attraction $\Basin{\param}{\variable}$ for each point on $\SolutionCurve$ (yellow).
      (b) Highlights the range of $\variableSpace$ in $\Basin{\paramMap{\hparam_1}}{\localMin_3}$ (purple) and the points on the homotopy curve from which $\Node{3}{1}$ is visible (red).
      (c) A sequence of nodes on $\SolutionCurve$ and their projections on the Basins of attraction of the next node in the sequence (dashed line).
      (d) Potential solution path found by \gls{pho} satisfying~\eqref{eqn:param_seq_requirements}.}
      \label{fig:pho}
\end{figure}

While nodes of the tree are elements of $\variableSpace\times\hparamSpace{\hparamSize}$, \gls{pho} only samples in $\hparamSpace{\hparamSize}$.
For a sampled $\hparam\in\hparamSpace{d}$, the $\localMin\in\variableSpace$ component of the node comes from solving the problem encoded by $\paramMap{\hparam}$.
This is a crucial feature of \gls{pho} since it allows the same problem $\NLP{\paramMap{\hparam}}$ to be solved multiple times, but with different initial guesses, as shown in Fig.~\ref{fig:pho}.
This leads to finding diverse solutions and better exploration of the trajectory space.

A summary of the \gls{pho} algorithm is given by Alg.~\ref{algo:pho} and~\ref{algo:sample_trial}.
Note that Alg.~\ref{algo:pho}, as well as the other algorithms, use the operator $\mathrm{sample}(\cdot)$ to randomly select an element from the set provided as an argument.
\gls{pho} begins by using the provided initial guess to solve the easy problem $\NLP{\paramMap{\mathbf{0}_\hparamSize}}$.
If this problem cannot be solved, \gls{pho} fails.
Thus, the easy problem should permit a large basin of attraction.
Next, the homotopy parameter set~$\hparamset$ and the solution attempts set~$\trials$ are initialized.
The homotopy parameter set keeps track of the candidate \gls{nlp}s, and the attempts set keeps track of which initial guesses we have attempted for each candidate \gls{nlp}.
The remainder of the algorithm involves swapping between the \textit{sampling} and \textit{solving} phases, depending on the solve/sample ratio.

The solve/sample ratio is computed via $\frac{|\trials|}{|\tree| \times |\hparamset|}$, where $\vert\cdot\vert$ gives the number of elements in a set.
Qualitatively, this number represents the fraction of possible unique $\mathrm{solve}()$ operations that have been attempted for the candidate \gls{nlp}s.
Thus, when $\frac{|\trials|}{|\tree| \times |\hparamset|}=1$, every candidate problem in $\hparamset$ has been attempted to be solved with every possible initial guess from $\tree$.
When the solve/sample ratio is below the threshold $\trialDensity$, \gls{pho} executes the solve phase, where it tries to add a node to $\tree$ by solving an existing problem in $\hparamset$ using an initial guess from $\tree$ that it has not yet tried on that problem.
If the result from the solver $\localMin_{new}$ is feasible and no similar trajectory exists, it will be added to $\tree$.
Regardless of the success of the solver, the attempt to solve $\NLP{\paramMap{\hparam}}$ will be added to $\trials$. 
Once enough solutions have been attempted to raise the ratio above the threshold, the sample phase is executed, where a new candidate parameterization is added to $\hparamset$.

\begin{algorithm}
\caption{Probabilistic Homotopy Optimization}
\begin{algorithmic}[1]
\Require Initial guess $\variable_0$
\Hyperparameters Iteration limit $q$, solve/sample threshold $\trialDensity$
\State $\tree \leftarrow \left\{ \left(\solve{\paramMap{\mathbf{0}_\hparamSize}}{\variable_0}, \mathbf{0}_\hparamSize \right)\right\}$
\State $\hparamset \leftarrow \{\mathbf{1}_\hparamSize,\mathbf{0}_\hparamSize\}$, \, $\trials \leftarrow \{\}$
\For{$q$ iterations}
    \If{$\frac{|\trials|}{|\tree| \times |\hparamset|} \le \trialDensity$}
        \State $(\Node{i}{j}, \hparam_\ell) = \mathrm{sampleAttempt}(\tree, \hparamset, \trials)$
        \State $\trials \leftarrow \trials \cup (\Node{i}{j}, \hparam_\ell)$
        \State $\localMin_{new} \leftarrow \solve{\paramMap{\hparam_\ell}}{\localMin_i}$
        \If{$\localMin_{new}\in\feasibleSet$ AND $\localMin_{new}\neq\localMin_j\,\forall \Node{j}{\cdot}\in\tree$}
            \State $\tree \leftarrow \tree \cup (\localMin_{new},\hparam_\ell)$
        \EndIf
    \Else
        \State $\hparamset \leftarrow \hparamset \cup \mathrm{sample}(\hparamSpace{d})$
    \EndIf
\EndFor
\State \Return $\{ \localMin_i: \Node{i}{j} \in \tree, \hparam_j = \mathbf{1}_\hparamSize \}$
\end{algorithmic} \label{algo:pho}
\end{algorithm}

\begin{algorithm}
\caption{sampleAttempt}
\begin{algorithmic}[1]
\Require Optimization Tree $\tree$, Parameters $\hparamset$, Attempts $\trials$
\Hyperparameters Probability of sampling goal $\goalProb$
\If{with probability $\goalProb$}
    \State $\mathbf{N} = \mathrm{sample}(\{ \Node{i}{j}\in\tree : (\Node{i}{j}, \mathbf{1}_\hparamSize)\notin\trials \})$
    \State \Return $(\mathbf{N}, \mathbf{1}_\hparamSize)$
\Else
    \State \Return $\mathrm{sample}(\{(\mathbf{N}\in\tree, \hparam\in\hparamset) : (\mathbf{N}, \hparam)\notin \trials \})$
\EndIf
\end{algorithmic}  \label{algo:sample_trial}
\end{algorithm}

\begin{figure}
    \centering
    \includegraphics[width=200pt]{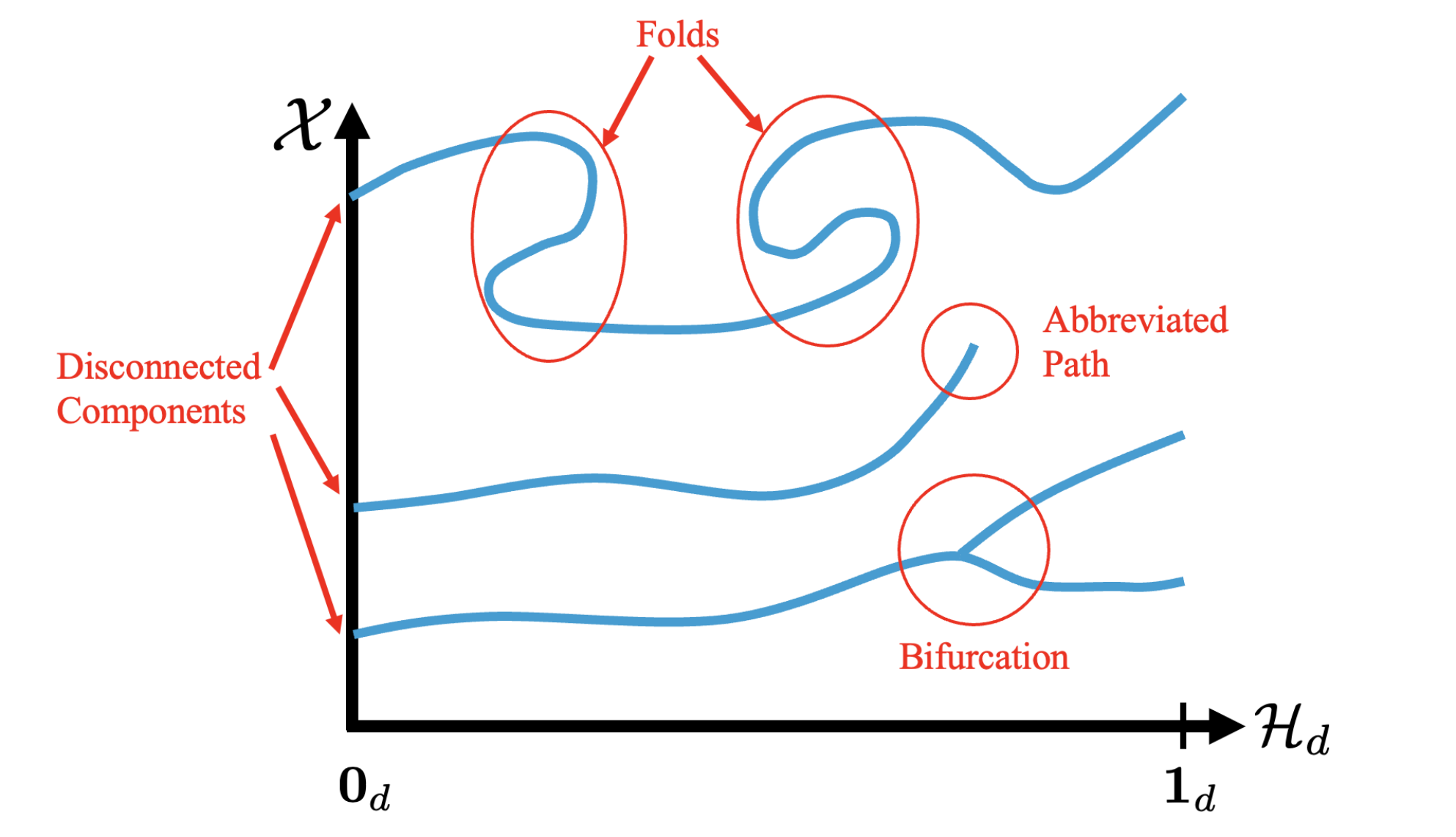}
    \caption{Visualization of common pitfalls encountered by Homotopy Methods.}
    \label{fig:pitfalls}
\end{figure}

\subsection{Algorithm Analysis}
Our algorithm can find solutions in the presence of common pitfalls encountered in homotopy methods.
These pitfalls are illustrated in Fig.~\ref{fig:pitfalls}.
While path tracing methods are restricted to continuously following a single connected component of the Solution Manifold, \gls{pho} can take discrete jumps, potentially large ones, in the homotopy parameter space.
These jumps can allow the \gls{pho} to skip over folds, move between disjoint components of the Solution Manifold, and continue progressing after an abbreviated path.
Furthermore, \gls{pho} supports storing multiple local minima for a given $\hparam$.
This allows the algorithm to traverse all branches emerging from a bifurcation and track multiple disjoint components of the Solution Manifold over the same region of the homotopy space.

If a sequence $\hparamSeq$ satisfying~\eqref{eqn:param_seq_requirements} exists, \gls{pho} will eventually sample a set of homotopy parameters that are close enough to $\hparamSeq$.
Since \gls{pho} tries every found solution as the initial guess to every candidate problem, the algorithm will eventually find a sequence of parameters to solve $\NLP{\paramMap{\mathbf{1}_\hparamSize}}$.
Note that this is not the same as guaranteeing that if a feasible solution $\NLP{\paramMap{\mathbf{1}_\hparamSize}}$ exists, \gls{pho} will find it.
The choice of parameters $\param$ for~\eqref{eqn:parameterized_nlp} and the homotopy map $\paramMap{\cdot}$ influence the number of iterations \gls{pho} needs to find a solution to $\NLP{\paramMap{\mathbf{1}_\hparamSize}}$.

\section{Linear Interpolation and RRT Homotopy Optimization} \label{sec:liho_and_rho}

For the sake of comparison, we discuss alternative methods for expanding nodes in the Optimization Tree to find a sequence $\hparamSeq$ that satisfies~\eqref{eqn:param_seq_requirements}.
In this section, we discuss \gls{lho} and \gls{rho} and compare them to \gls{pho}.

\subsection{Linear Interpolation Homotopy Optimization}

While \gls{pho} supports $\hparam$ of arbitrary dimension, \gls{lho} requires a scalar homotopy parameter $\hparam\in\hparamSpace{1}$ and uses the parameter mapping 
\[
\paramMap{\hparam} = (1-\hparam)\easyParam + \hparam\goalParam.
\]
This scalar parameterization produces an algorithm similar to the original Homotopy Optimization Method proposed in~\cite{Dunlavy2005HomotopyOM}.
They are similar in that both use a one-dimensional homotopy parameter to transform a simple optimization (rather than a system of equations) into a difficult one.
However, \gls{lho} uses a dynamic step adjusting strategy for $\hparam$, while the original Homotopy Optimization Method uses a predetermined, fixed step size.
Specifically, \gls{lho} uses the following strategy
\begin{itemize}
    \item After each $k_1$ consecutive rounds of optimization being solvable $\Delta\hparam$ is multiplied by a constant $c_1 > 1$
    \item after $k_2$ consecutive failures $\Delta \hparam$ is multiplied by $c_2 < 1$
    \item Terminate algorithm if $\Delta\hparam < \epsilon$ occurs
\end{itemize}
where $k_1$, $c_1$, $k_2$, $c_2$, $\epsilon$, and the initial $\Delta\hparam$ are all hyperparameters.
This strategy is useful as it avoids predetermining the steps and allows the algorithm to increase the step size quickly in "easy" regimes. 
A summary of the \gls{lho} algorithm is shown in Algorithm~\ref{algo:liho}.

\begin{algorithm} 
\caption{Linear Interpolation Homotopy Optimization}
\begin{algorithmic}[1]
\Require Initial guess $\variable_0$
\Hyperparameters $k_1, c_1, k_2, c_2, \epsilon, \Delta\hparam_0$
\State $\localMin_{cur} \leftarrow \solve{\paramMap{\mathbf{0}_\hparamSize}}{\variable_0}$
\State $\hparam \leftarrow \mathbf{0}_d$, \, $\Delta\hparam \leftarrow \Delta\hparam_0$
\While{$\hparam \neq \mathbf{1}_\hparamSize$}
    \State $\hparam \leftarrow \hparam + \Delta\hparam$
    \State $\localMin_{new} = \solve{\paramMap{\hparam}}{\localMin_{cur}}$
    \If{$\localMin_{new}\in\feasibleSet$}
        \State $\localMin_{cur} \leftarrow \localMin_{new}$
        \State Re-adjust $\Delta\hparam$
    \Else
        \State Re-adjust $\Delta\hparam$ or \Return $NULL$
    \EndIf
\EndWhile
\State \Return $\localMin_{cur}$
\end{algorithmic} \label{algo:liho}
\end{algorithm}

\subsection{RRT Homotopy Optimization}
\gls{rho} is inspired by another probabilstic search algorithm, \gls{rrt}~\cite{lavalle2001rapidly}.
The key distinction holds: \gls{rho} samples only in the homotopy parameter space $\hparamSpace{\hparamSize}$, not the full configuration space associated with the nodes of the tree.
\gls{rho} grows the Optimization Tree by sampling a point $\hparam_{new}\in\hparamSpace{\hparamSize}$, finding the node $\closeNode\in\tree$ with the closest homotopy parameter to the sampled point, and then attempting to connect the sampled point to $\closeNode$ by solving $\NLP{\paramMap{\hparam_{new}}}$ using  $\localMin_{close}$ for an initial guess.
A summary of the \gls{rho} algorithm is shown in Algorithm~\ref{algo:rho}.

\begin{algorithm} 
\caption{RRT Homotopy Optimization}
\begin{algorithmic}[1]
\Require Initial guess $\variable_0$
\Hyperparameters Goal probability $\goalProb$, iteration limit $q$
\State $\tree \leftarrow \left\{ \left(\solve{\paramMap{\mathbf{0}_\hparamSize}}{\variable_0}, \mathbf{0}_\hparamSize \right)\right\}$
\For{$q$ iterations}
    \If{with probability $\goalProb$}
        \State $\hparam_{new} \leftarrow \mathbf{1}_\hparamSize$
    \Else
        \State $\hparam_{new} \leftarrow \mathrm{sample}(\hparamSpace{\hparamSize})$
    \EndIf
    \State $\closeNode \leftarrow \arg\min_{\Node{i}{j}\in\tree} \Vert \hparam_{new} - \hparam_j \Vert$
    \State $\localMin_{new} \leftarrow \solve{\paramMap{\hparam_{new}}}{\localMin_{close}}$
    \If{$\localMin_{new}\in\feasibleSet$}
        \State $\tree\leftarrow\tree\cup (\localMin_{new}, \hparam_{new})$
        \If{$\hparam_{new} = \mathbf{1}_\hparamSize$}
            \State \Return $\localMin_{new}$
        \EndIf
    \EndIf
\EndFor
\State \Return $NULL$
\end{algorithmic} \label{algo:rho}
\end{algorithm}

Important implementation details for \gls{rho} include the sampling distribution and the distance metric used to determine the closest node in the tree.
We chose the sampling distribution to be uniform over $\hparamSpace{\hparamSize}$ (Line 6) and we use Euclidean distance metric in parameter space (Line 8).

\begin{figure}
    \centering
    \includegraphics[width=\columnwidth]{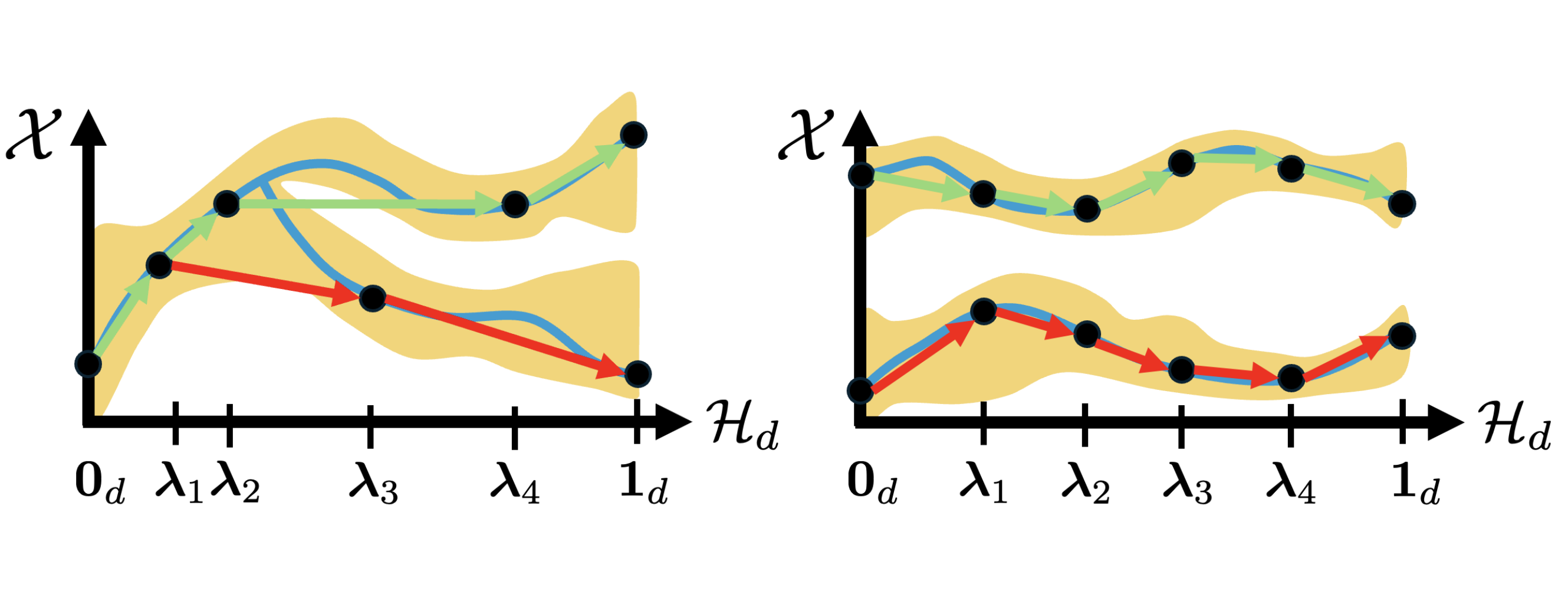}
    \caption{Illustration of how \gls{rho} and \gls{pho} deal with bifurcation (left) and multiple local minima (right). After the green path has been found, only \gls{pho} can also find the red path.}
    \label{fig:pho_comparison}
\end{figure}

\subsection{Comparison with P-HO} \label{ssec:theoretical_comparison}

Although we observed \gls{rho}, and to a lesser extent \gls{lho}, to be effective algorithms, we also observed that both of them are susceptible to converge to lower quality local minima compared to \gls{pho}.
An analysis of the Solution Manifolds can explain this.
Analyzing the Solution Manifolds of real-world systems, even toy systems like the cart-pole, is intractable.
Therefore, we first discuss how the structure of Solution Manifolds impacts our presented algorithms using visuals of hypothetical Solution Manifolds.
Later in Sec.~\ref{sec:results}, we will present empirical results supporting this theoretical analysis.

\subsubsection{Bifurcation}

In case of bifurcation of the Solution Manifold, \gls{rho} and \gls{lho} can only follow one of the branches.
This is because once a node on one branch is chosen, that node will be the closest neighbor any time a parameter in the bifurcated region is sampled. 
Because \gls{pho} allows multiple nodes to have the same parameter and does not expand the tree based on the proximity of sampled nodes, it can explore multiple branches.
This distinction between the algorithms is illustrated in Fig.~\ref{fig:pho_comparison}.

\subsubsection{Distance Bias}

\gls{rho} assumes that if $\hparam_i$ is the closest to $\hparam_j$, then the solution found for $\NLP{\paramMap{\hparam_i}}$ will have the best chance of being in a Basin of Attraction associated with $\NLP{\paramMap{\hparam_j}}$.
This assumption, however, is fundamentally flawed. 
First of all, it is predicated on the definition of ``closeness" between two points in a space where distance can be ill-defined.
\gls{pho} requires no such definition.
Furthermore, consider the case where a local minima near the goal is found and is not in the basin of attraction of the goal solution. 
\gls{rho} may fail to converge since all samples in the goal region will find this node as the closest node.

\subsubsection{Diversity of Solutions}
Since \gls{pho} maintains multiple different solutions for each $\hparam$, it can explore different local minima for a single $\hparam$.
This results in a diverse set of solutions for both the end result and the intermediary nodes. 
Consider, for example, Fig.~\ref{fig:pho_comparison}, where the Solution Manifold consists of multiple disjoint components.
Because \gls{rho} cannot store multiple solutions $\localMin$ for a given $\hparam$, it is restricted to following either the top component of the manifold or the bottom component.
Since the top and bottom components find different local minima for the goal problem, the quality of the solution \gls{rho} returns is stochastic.
\gls{pho} can store multiple solutions per $\hparam$, so it will find the local minima for both the top and bottom components.

\section{Results} \label{sec:results}

We supplement the theoretical analysis of \gls{pho} from Sec.~\ref{ssec:theoretical_comparison} with empirical results from case studies involving dynamic motion planning for two systems: the underactuated Cart-Pole and the MIT Humanoid robot.
The hyperparameters used for all algorithms for both case studies are given by Table~\ref{tab:hyperparams}, unless noted otherwise.

\begin{table}[h!]
\centering
\caption{Hyperparameters for Homotopy Optimization Algorithms}
\label{tab:hyperparams}
\resizebox{\columnwidth}{!}{%
\begin{tabular}{|c|c|c|c|c|c|c|c|c|}
\hline
$\goalProb$ & $\trialDensity$ & $q$ & $k_1$ & $k_2$ & $c_1$ & $c_2$ & $\epsilon$ & $\Delta\hparam_0$ \\
0.3         & 1.0             & $10^4$ & 2     & 1     & 1.5   & 0.3   & $10^{-9}$       & $10^{-2}$ \\         
\hline
\end{tabular}%
}
\end{table}

\subsection{Cart-Pole}

Cart-Pole is a simple yet dynamically rich system often studied in robotics. 
The motion planning task is to swing up the pole by only applying horizontal force to the cart.
We benchmark \gls{pho}, \gls{rho}, \gls{lho} algorithms on this task and compare them to directly optimizing from $\Node{0}{0}$ as an initial guess (referred to as one-depth method).

\begin{figure}[!thb]
    \centering
    \includegraphics[width=.45\linewidth]{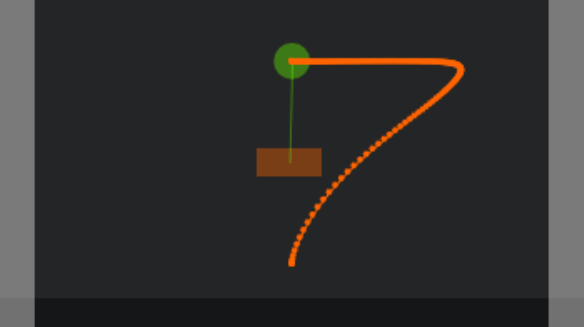}
    \includegraphics[width=.45\linewidth]{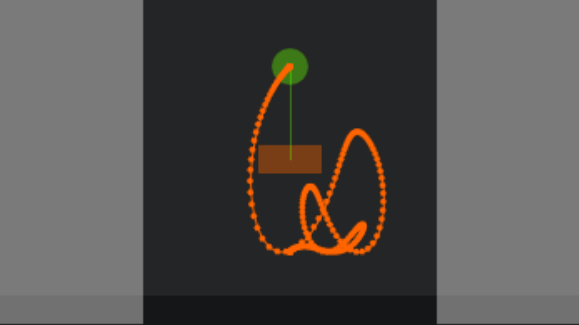}
    \caption{The trajectories $\localMin_0, \localMin_K$ found by \gls{pho} shown from left to right.
    As we go from $\param_0$ to $\param_K$, $m_{pole}$ increases from $1kg$ to $60kg$ and $F_{max}$ goes from $200N$ to $100N$ thus more swings are needed.
    }    
    \label{fig:cartpole-param-evolution}
\end{figure}

There homotopy parameters are: mass of the cart ($m_{cart}$), mass of the pole ($m_{pole}$), force limit ($F_{max}$), length of the pole ($l_{pole}$), and horizontal movement range of the cart ($x_{max}$).
The time horizon is 5 seconds.
Figure~\ref{fig:cartpole-param-evolution} shows an example of how the trajectory evolves as parameters change.

\begin{figure}[thb]
    \centering
    \includegraphics[width=1\linewidth]{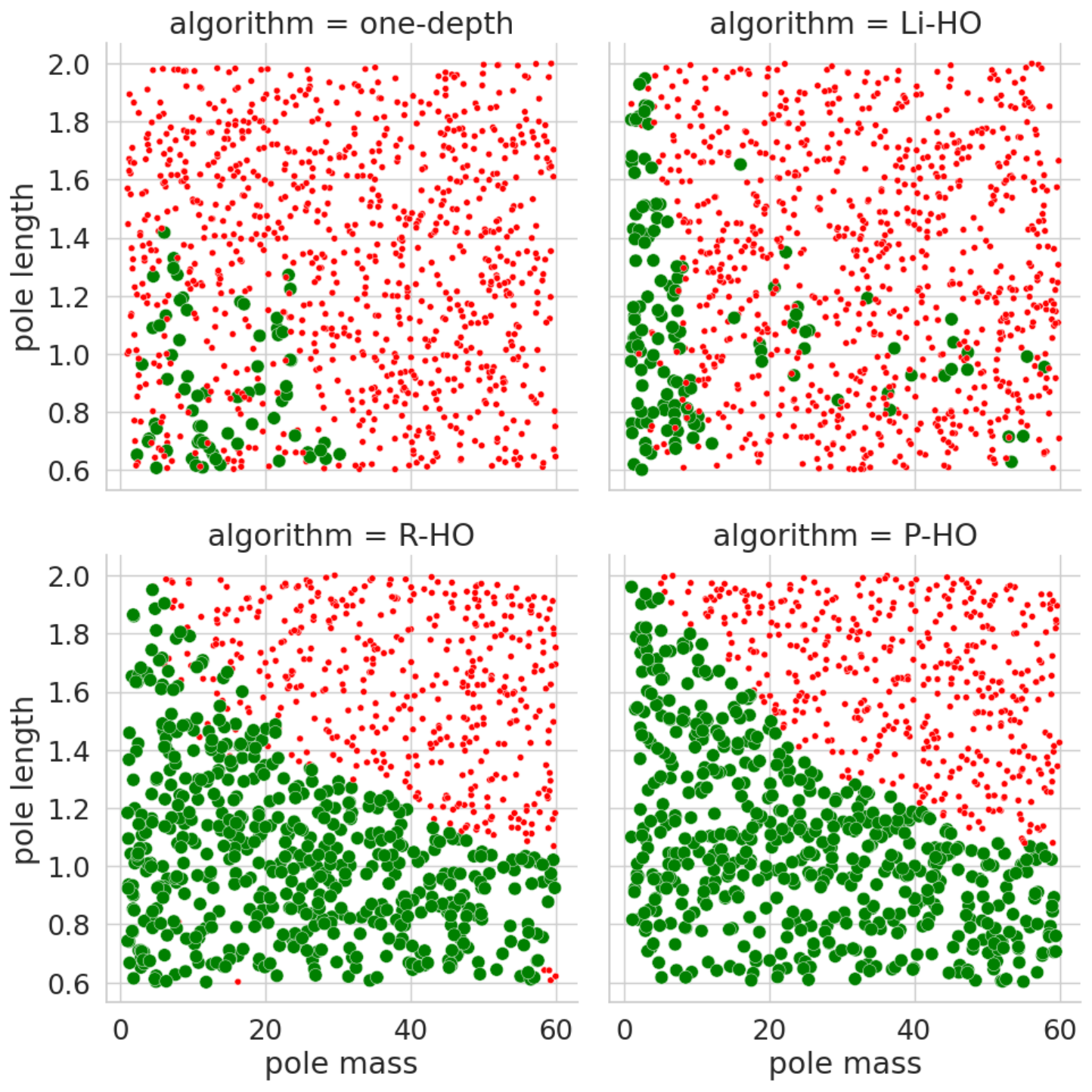}
    \caption{Result of solving Cart-Pole swing up. For each parameter $\param$, a green dot represents the algorithm solving $\NLP{\param}$ successfully within the iteration limit, and a red dot represents an unsuccessful attempt. Parameters are uniformly drawn from $1kg \leq m_{pole} \leq 60kg$, $0.6m \leq l_{pole} \leq 2m$, $x_{max}=1.6m$, $F_{max}=100N$, $m_{cart}=20kg$ (SI units). With an initial guess of zero, none of the $\NLP{\param}$ in this range are solvable. 
    }
    \label{fig:cartpole-exploring-algorithms}
\end{figure}

To compare the performance of different algorithms, 1000 samples of $\param_i\in\paramSpace$ are uniformly selected from a "difficult" region in parameter space.
Starting from the same root node $\Node{0}{0}$, each algorithm attempts to solve for $\param^*=\param_i$ using no more than 200 queries of the NLP solver.
The results in Fig.~\ref{fig:cartpole-exploring-algorithms} show \gls{rho} and \gls{pho} outperform \gls{lho} and the one-depth method.
Additionally, \gls{lho} is extremely sensitive to small changes in the goal parameters as we observe the set of solvable goal parameters using \gls{lho} is extremely sparse.
Such sparsity may seem non-intuitive, but it can be explained by the potential presence of bifurcations, discontinuities, and folds in the Solution Manifold.
In general, the path that \gls{lho} tries to traverse may contain regions of physical infeasibility.
\gls{rho} and \gls{pho} avoid these by finding a path in higher dimensions.

\subsection{MIT Humanoid}

\begin{figure}
    \centering
    \includegraphics[width=\columnwidth]{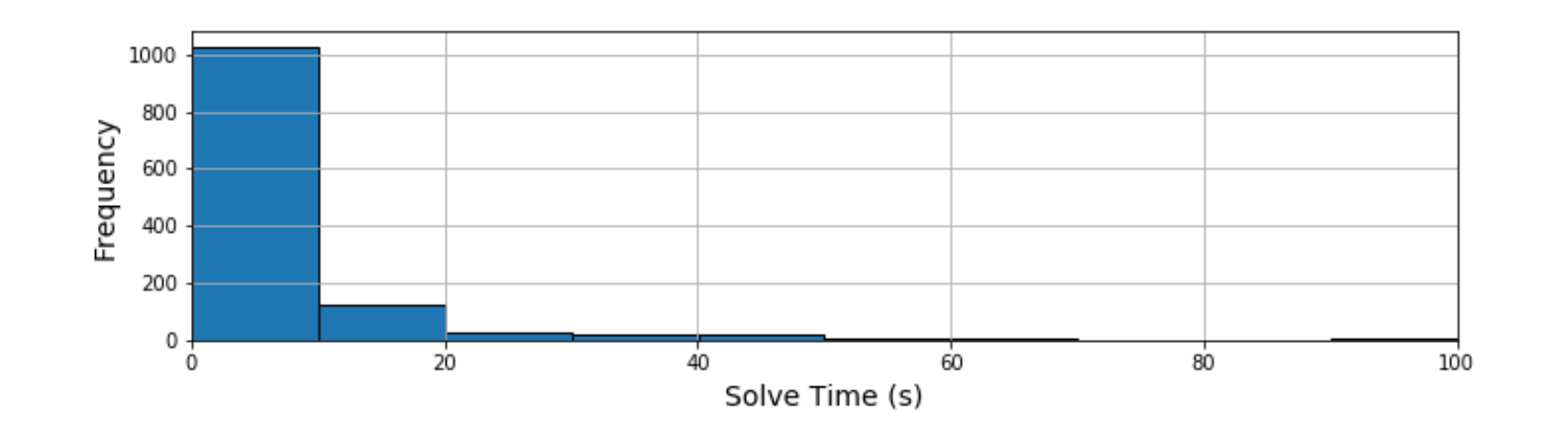}
    \caption{Histogram of solve times for the MIT Humanoid back/front flip trajectory optimizations solved over the course of \gls{lho}, \gls{rho}, and \gls{pho}.}
    \label{fig:solve_times}
\end{figure}

Next, we will look at dynamic motion planning for the MIT Humanoid.
Details about the baseline formulation can be found in~\cite{chignoli2021humanoid}.
The baseline motion planning problem was solved using KNITRO's interior point algorithm~\cite{byrd2006k} but required substantial cost function tuning and manual initial-guess crafting.
For the current work, we modify the baseline formulation in ways that make the original, manual solution method intractable, thus necessitating the algorithm proposed in this work.
The key features of the modified formulation are:
\begin{itemize}
    \item Modeling the actuator level dynamics of the robot, including the kinematic loops~\cite{chignoli2023recursive} that arise,
    \item Constraining the power consumed by the robot's motors via a quadratic regression of their efficiencies,
    \item Planning up to the point of landing (rather than the point of takeoff).
\end{itemize}
The typical solve times for this trajectory optimization problem are shown in Fig.~\ref{fig:solve_times}.

\begin{figure}[thb]
    \centering
    \includegraphics[width=\columnwidth]{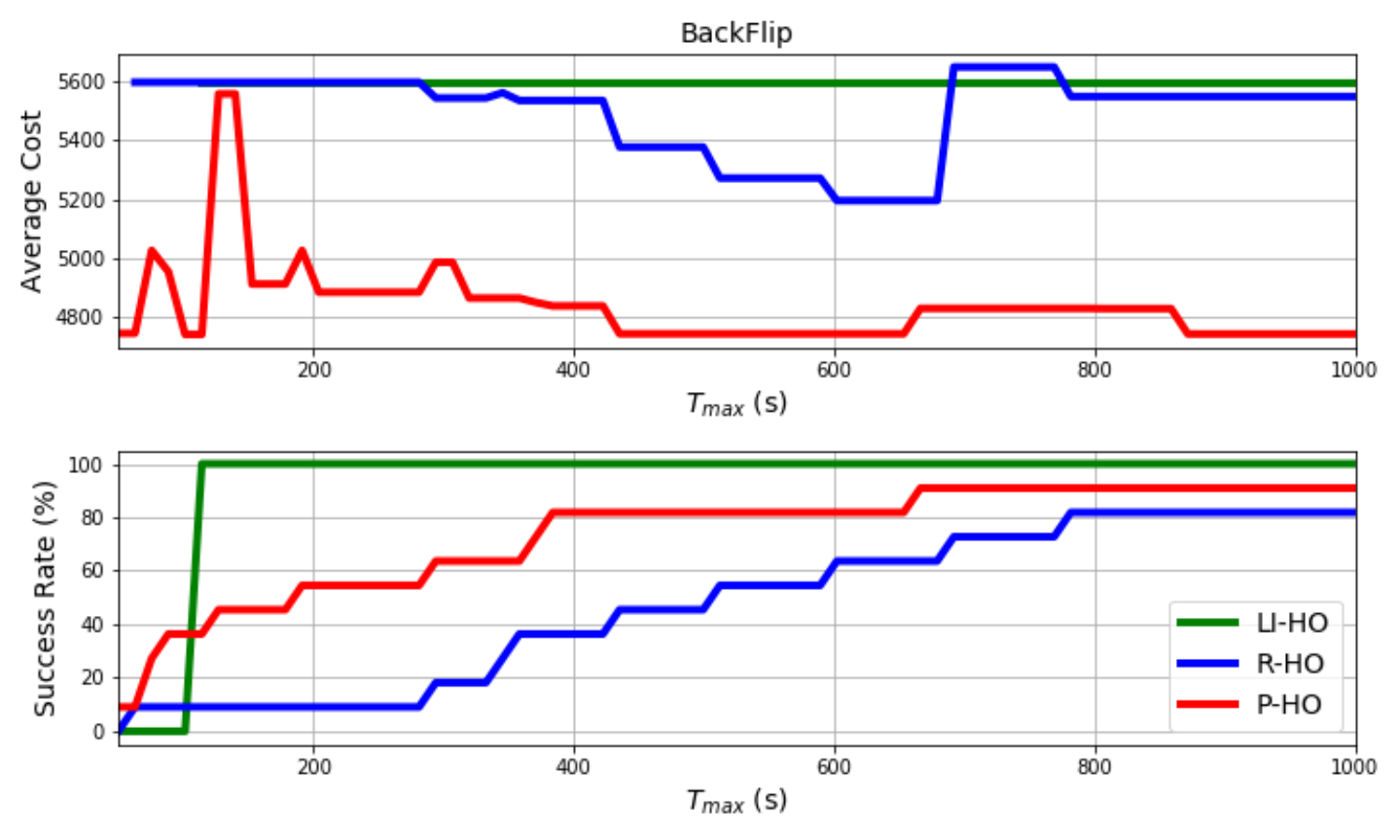}
    \caption{Average cost and success rate for planning a back flip of the MIT Humanoid robot within $\timeLimit$.}    \label{fig:back_flip_solve}
\end{figure}

\begin{figure}[thb]
    \centering
    \includegraphics[width=\columnwidth]{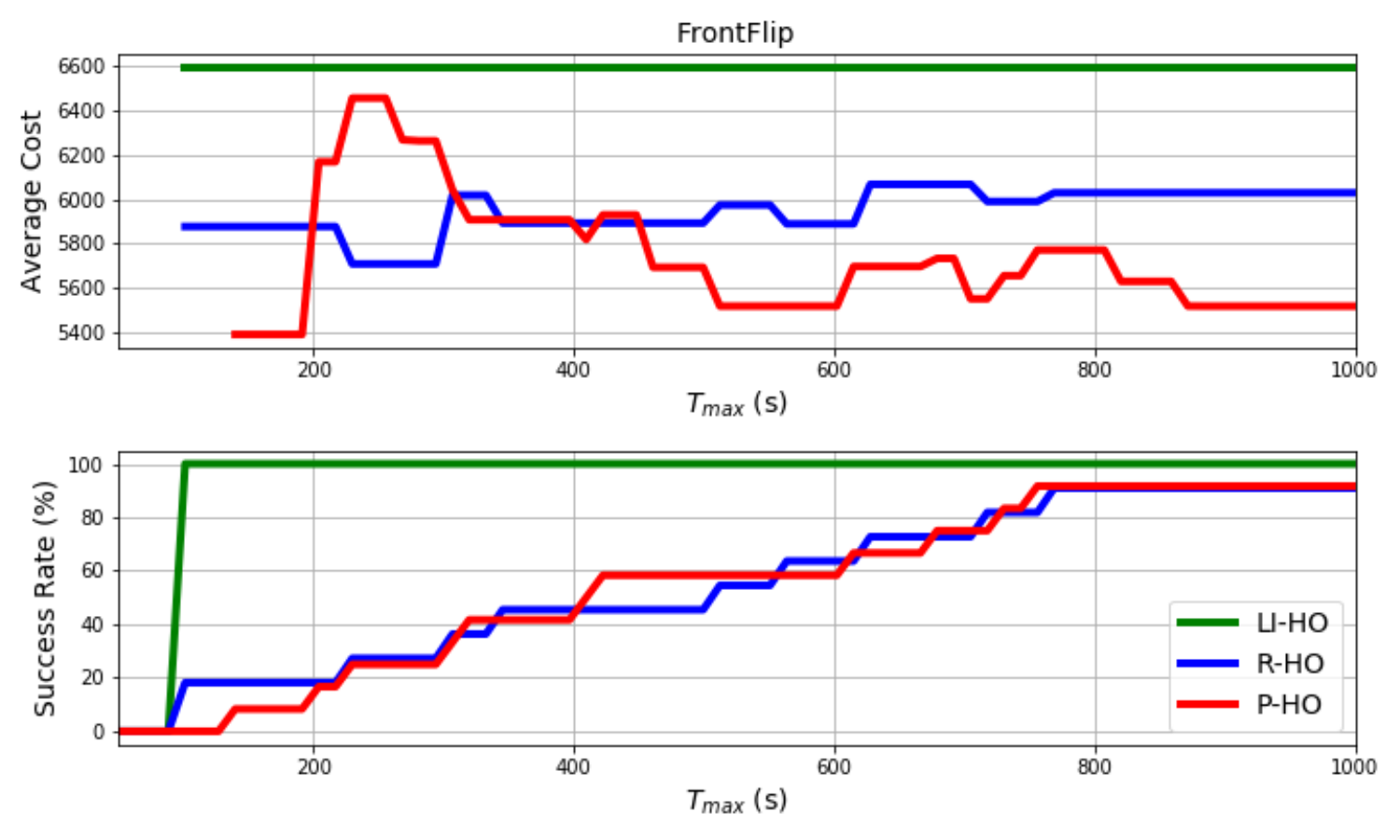}
    \caption{Average cost and success rate for planning a front flip of the MIT Humanoid robot within $\timeLimit$.}
    \label{fig:front_flip_solve}
\end{figure}

We compare the performance of the \gls{lho}, \gls{rho}, and \gls{pho} with the following experiment.
For the sake of fairness, we set a total solve-time limit $\timeLimit$ on each algorithm.
For each algorithm, if the desired motion planning problem is solved prior to $\timeLimit$, the objective function value of the solution corresponding to $\goalParam$ and the total solve time required are recorded.
Thus, we compare the algorithms based on two criteria: (1) average cost when solved within $\timeLimit$ and (2) likelihood to find a solution within $\timeLimit$, referred to as ``success rate."
\gls{lho} is deterministic, its success rate within a given time interval is either $0\%$ or $100\%$, and the cost associated with the solution it finds every time is fixed.
\gls{rho} and \gls{pho}, howevever are probabilistic.
Therefore, these algorithms were repeated and averaged over $\numExperiments$ runs per motion planning problem.
The results of the comparisons for a back flip~(Fig.~\ref{fig:front_flip}) and front flip planning problem are shown in Fig.~\ref{fig:back_flip_solve} and~\ref{fig:front_flip_solve}, respectively.
The parameters that define these motion planning homotopies, as well as their values, are shown in Table~\ref{tab:flip_parameters}.

\begin{table}[htbp]
\caption{Homotopy Parameters for Humanoid Motion Planning}
\label{tab:flip_parameters}
\resizebox{\columnwidth}{!}{%
\begin{tabular}{r|cc|cc|}
\cline{2-5}
\multicolumn{1}{c|}{\multirow{2}{*}{}} & \multicolumn{2}{c|}{\textbf{Back Flip}} & \multicolumn{2}{c|}{\textbf{Front Flip}} \\ \cline{2-5} 
\multicolumn{1}{c|}{} & \multicolumn{1}{c|}{\textbf{$\easyParam$}} & \textbf{$\goalParam$} & \multicolumn{1}{c|}{\textbf{$\easyParam$}} & \textbf{$\goalParam$} \\ \hline
\multicolumn{1}{|r|}{\textbf{Dynamics Timestep (s)}} & \multicolumn{1}{c|}{0.03} & 0.01 & \multicolumn{1}{c|}{0.03} & 0.01 \\ \hline
\multicolumn{1}{|r|}{\textbf{Mid-Flight Body Height (m)}} & \multicolumn{1}{c|}{0.4} & 1.2 & \multicolumn{1}{c|}{0.4} & 1.2 \\ \hline
\multicolumn{1}{|r|}{\textbf{Final Pitch Rotation ($^\circ$)}} & \multicolumn{1}{c|}{0} & -360 & \multicolumn{1}{c|}{0} & 360 \\ \hline
\multicolumn{1}{|r|}{\textbf{Self-Collision Relaxation}} & \multicolumn{1}{c|}{50} & 1 & \multicolumn{1}{c|}{50} & 1 \\ \hline
\end{tabular}%
}
\end{table}

We first note that Homotopy Optimization is required for these motion planning problems.
Neither can be solved with a trivial initial guess, such as all zeros or a constant joint position at every timestep.
For both motions, \gls{lho} is the fastest at finding a valid solution to the desired problem in the sense that \gls{lho} always finds solutions within $100s$, while \gls{rho} and \gls{pho} take at least $700s$ to reach success rates of $80\%$ or better.
However, also, in both cases, the solution \gls{lho} finds has a significantly higher objective function value than the solutions returned by \gls{rho} and \gls{pho}.
For these problems, the objective function value corresponds to control effort, so finding high-quality (i.e., low-cost) local minima is crucial.
Thus, if the user is willing to sacrifice time efficiency for solution quality, \gls{rho} and \gls{pho} are better options.

Lastly, \gls{pho} has the advantage over \gls{rho} by constantly improving its solution until $\timeLimit$ is reached.
Therefore, as time passes, \gls{pho} is likely to find better local minima, even the global minimum, if such a path exists.
\gls{rho}, on the other hand, terminates once a solution is found.
While ``anytime" variants such as RRT* exist~\cite{karaman2011anytime}, they rely on rewiring the tree as new nodes are added based on the edge costs.
However, our search problem does not have edge costs.
The cost for our problem is the objective function value of the solved goal \gls{nlp}, which is only accrued once the goal node has been reached.
This incompatibility with ``anytime" variants can explain the unintuitive increase over time shown by the average cost of \gls{rho}.
Analyzing the structure of the Solution Manifold for such a complex problem is outside the scope of this work. 
However, we hypothesize that a bifurcation in the Solution Manifold could explain the increasing cost. 
While \gls{pho} could follow all branches emerging from the bifurcation, \gls{rho} can only follow one branch, and the choice of the branch can be stochastic.
One branch may efficiently lead to a low-cost solution while the other leads slowly to a high-cost solution.

\section{Conclusion} \label{sec:conclusion}

This work presented the \gls{pho} algorithm, a homotopic approach for solving challenging, optimization-based motion planning problems.
The algorithm emerges from formulating the problem of multidimensional Homotopy Optimization as a search problem in the space of homotopy parameters.
Given this formulation, we described the \gls{pho} algorithm and analyzed its properties, such as its ability to deal with bifurcated, disconnected, and folded Solution Manifolds.
\gls{pho} was then compared to alternative Homotopy Optimization methods, the deterministic \gls{lho} and the stochastic \gls{rho}.
Two case studies were presented. The Cart-Pole study showed that \gls{pho} and \gls{rho} methods reliably solve much more challenging optimization tasks. The MIT Humanoid study showed that \gls{pho} is consistently able to find higher-quality local minima if some additional time is allowed for the algorithm to run.

Future work will focus on leveraging knowledge of the parameter space to make search more efficient, as well as investigating alternative parameterizations.
For example, finding a meaningful distance metric in the parameter space can improve the notion of ``closest node" in allow \gls{rho} and lead to more efficient tree growth.
For \gls{pho}, non-uniform parameter sampling strategies, such as Gaussian Sampling in \gls{prm} \cite{772447}, can lead to a more efficient search.
Finally, we will extend the algorithm to enable discrete parameterizations that will allow, for example, constraints to be altogether turned on and off rather than simply softened.

\section*{Acknowledgments}
This work was supported by Naver Labs, LG Electronics, and the National Science Foundation (NSF) Graduate Research Fellowship Program under Grant No. 4000092301.

\bibliographystyle{ieeetr}
\bibliography{references}

\vfill

\end{document}